# Playful AI in Professional Email: A Field Experiment on Tone and Recipient Engagement

Ziv Ben-Zion[1,2,*] and Teddy Lazebnik[3,4,x,*]

[1] School of Public Health, Faculty of Social Welfare and Health Sciences, University of Haifa, Haifa, Israel

[2] Department of Psychiatry, Yale School of Medicine, New Haven, CT, USA

[3] Department of Information Systems, University of Haifa, Haifa, Israel

[4] Department of Computing, Jönköping University, Jönköping, Sweden

*These authors contributed equally

[x] Corresponding author: teddy.lazebnik@ju.se

## Abstract

Large language models (LLMs) are rapidly reshaping workplace communication, yet whether AI-assisted writing changes how recipients actually behave, and through what channel, remains unknown. Here, in a randomized crossover field experiment, 121 employees across six companies sent work emails under three conditions over three weeks: unaided writing, GPT-5 rewriting in a playful tone, and GPT-5 rewriting in a professional tone. Across 16,880 emails, playful editing increased emotional positivity ($B=+0.068$, $p<0.001$), and professional editing decreased it ($B=-0.041$, $p<0.001$), yet neither condition directly altered open rates, reply rates, or response times. Instead, within-sender positivity strongly predicted both opening ($OR=2.05$) and replying ($OR=3.32$, $p<0.001$), a significant indirect pathway through which AI editing shaped behavior, in the absence of any direct effect. These findings suggest that AI-assisted communication shapes workplace engagement not through its use, but through the emotional tone of the language it produces.

## Introduction

Generative artificial intelligence (AI) has moved, in only a few years, from research curiosity to an everyday co-author of workplace communication. A majority of knowledge workers now use AI at work[1], and drafting and polishing written email messages is among its established everyday uses[2]. Because such tools measurably improve the quality and efficiency of written output[3,4], organizations adopting them assume that a better-written message is a more effective one: that improving the tone or polish of an email will make recipients more likely to read it and reply. Nevertheless, this assumption has never been tested where it matters - in the volume and messiness of real organizational email, with the people who receive it.

Whether AI-assisted writing changes recipient behavior is, at its core, a question about mechanism. Decades of research in organizational behavior and social psychology show that the emotional texture of a message (e.g., its warmth, playfulness, positivity) strengthens rapport, reduces perceived social distance, and invites reciprocal engagement[5,6]. If Large language models (LLMs) editing affects recipients at all, it may do so not because a message is "AI-assisted", but because it shifts these affective features of language. The distinction is not merely semantic. The AI-mediated communication framework holds that AI shapes interpersonal outcomes through the features of the messages it produces, not through its mere presence[7]. Recent work has begun to trace such indirect pathways, for instance, showing that repeated human–AI interaction can amplify users' own biases without their awareness[8] and that AI-generated language can out-persuade human language when personalized[9].

LLMs are also a uniquely tractable tool for isolating tone from content. Due to the fact that they can rewrite a human draft while preserving its meaning, they allow the substance of a message to remain fixed while randomizing its stylistic register. This manipulation is difficult to achieve at scale with human writers, but it is a natural affordance of instruction-tuned models. The emotional sensitivity we and others have documented in LLMs' own outputs and behavior[10–12] becomes an experimental instrument. Namely, a way to ask not only *whether* AI editing changes workplace engagement, but *through what channel*.

In this study, we report a randomized crossover field experiment in which 121 employees across six companies sent their ordinary work emails under three conditions: (1) unaided writing ('Unaided'), (2) playful LLM rewriting ('Playful'), and (3) professional LLM rewriting ('Professional'), in counterbalanced order over three weeks. Since every sender served as their own control across all three conditions, the design isolates the within-person effect of tone while eliminating stable individual differences in writing style or recipient network. Outcomes were the behaviors that matter to a sender (i.e., whether each email was opened, whether it was answered, and how quickly) together with an automated measure of the emotional positivity of every outgoing message.

We hypothesized that playful editing would increase engagement across the board (i.e., more opens, more replies, faster responses) compared to unaided writing, consistent with the view that positive affect and playfulness signal approachability and motivate reciprocation[5,6,13], and that professional editing would dampen it by producing a more generic, less personally distinctive voice[14].

Our results did not support this primary hypothesis. Neither condition changed opening, replying, or response speed. In an exploratory analysis, however, a consistent pattern emerged: playful editing reliably increased the emotional positivity of emails, professional editing decreased it, and positivity (measured email-by-email within each sender) strongly predicted whether a message was opened and answered. The behavioral effect of AI editing ran entirely through tone - a significant indirect pathway, even though editing had no direct effect on behavior. The LLM was neither a bonus nor a penalty in itself; it mattered only insofar as it changed the emotional character of the language that recipients read.

## Results

### Sample characteristics and descriptive patterns

One hundred and twenty-one (121) employees from six companies participated, sending a total of 16,880 emails across the three experimental conditions (Unaided: n=5,633; Professional: n=5,595; Playful: n=5,652). The sample comprised 88 men (72.7%) and 33 women (27.3%), with a mean age of 38.5 years (SD=7.4; range 22–57). Companies represented the software, hardware, services, and medical-device sectors across four countries (Israel, the United Kingdom, Sweden, and Ukraine; Table S1). Internal emails (sent to colleagues within the same company) accounted for 41.4% of the total, while external emails accounted for 58.6%.

Rates of email opening and replying were broadly similar across conditions (Table 1): Unaided (57.0% opened, 22.2% replied), Professional (58.0% opened, 21.5% replied), and Playful (57.7% opened, 22.2% replied). Median time to first open was approximately 25 hours across all conditions; median time to first reply was approximately 38 hours. Mean positivity scores (an automated sentiment-polarity score from −1 to +1 capturing each message's emotional tone) differed visibly: Unaided (0.072), Professional (0.028), Playful (0.136), suggesting that LLM-assisted editing shifted the emotional register of outgoing messages in the expected directions.

| Condition | Number of emails | Email open rate | Email reply rate | Positivity (mean ± SD) | Median time to open (h) | Median time to reply (h) |
|---|---|---|---|---|---|---|
| Unaided | 5,633 | 57.0% | 22.2% | 0.072 ± 0.42 | 25.1 | 38.1 |
| Professional | 5,595 | 58.0% | 21.5% | 0.028 ± 0.42 | 25.9 | 38.7 |
| Playful | 5,652 | 57.7% | 22.2% | 0.136 ± 0.41 | 24.7 | 38.8 |

**Table 1. Descriptive statistics by condition.** Conditions: Unaided (participant's own draft), Professional (LLM rewrite in a professional register), and Playful (LLM rewrite in a playful register). Positivity is an automated sentiment-polarity score ranging from −1 (most negative) to +1 (most positive), shown as mean ± SD. Open and reply rates are the percentages of emails opened and replied to; median times to first open and first reply are given in hours.

### Condition effects on email positivity (a-path)

A linear mixed-effects model with random intercepts per sender and fixed effects for condition confirmed a significant omnibus effect of condition on positivity score ($\chi^2(2)=206$, $p<0.001$; Fig. 1a). Playful LLM editing increased positivity relative to unaided writing ($B=0.068$, $SE=0.008$, $p<0.001$), and professional-style editing decreased it ($B=-0.041$, $SE=0.008$, $p<0.001$). All pairwise differences were significant after Holm correction, with the largest contrast between Playful and Professional ($B=0.109$, $p<0.001$).

The effect of condition on positivity interacted with recipient type ($\chi^2(2)=24.0$, $p<0.001$). The positivity advantage of Playful over Professional was stronger for external recipients (0.134 units, $p<0.001$) than for internal recipients (0.066 units, $p<0.001$). This asymmetry reflects a lower baseline positivity in external emails, which were markedly less positive than internal emails even without AI assistance (Unaided condition: $M=-0.05$ vs $M=0.24$, respectively; $t=27.3$, $p<0.001$); their more neutral baseline register leaves a larger dynamic range for LLM-mediated tone shifts.

### Absence of direct condition effects on behavioral outcomes (c-path)

Generalized linear mixed-effects models (binomial family, random intercepts per sender) revealed no significant omnibus effect of condition on open rate ($\chi^2(2)=0.94$, $p=0.62$) or reply rate ($\chi^2(2)=2.34$, $p=0.31$). Pairwise contrasts were uniformly non-significant after Holm correction (all $p>0.50$). Cox proportional hazards models stratified by sender — a within-subject survival model - similarly showed no condition effect on time to first open ($\chi^2(2)=1.69$, $p=0.43$) or time to first reply ($\chi^2(2)=2.83$, $p=0.24$; Table 2). Hazard ratios for all pairwise comparisons fell close to the null value of 1.0 (range 0.94-1.03) with wide confidence intervals spanning 1.0. These null findings were consistent across internal and external recipient subgroups.

| Outcome | Model | Test statistic | p |
|---|---|---|---|
| Emails opened | GLMM (binomial) | $\chi^2(2)=0.94$ | 0.62 |
| Emails replied | GLMM (binomial) | $\chi^2(2)=2.34$ | 0.31 |
| Positivity score | LMM | $\chi^2(2)=206$ | <0.001* |
| Time to open | Cox (stratified) | $\chi^2(2)=1.69$ | 0.43 |
| Time to reply | Cox (stratified) | $\chi^2(2)=2.83$ | 0.24 |

**Table 2. Omnibus condition effects on all outcomes.** Each row reports the omnibus test of experimental condition (Unaided, Professional, Playful) on one outcome. LMM, linear mixed-effects model with random intercepts per sender; GLMM, generalized linear mixed-effects model (binomial family); Cox (stratified), Cox proportional-hazards model stratified by sender. Test statistics are likelihood-ratio $\chi^2$ (df = 2); p values are for the omnibus condition effect. *marks the only significant effect (email positivity).

### Positivity predicts behavioral engagement within senders (b-path)

To test whether email-level positivity scores predicted opening and replying — independent of condition — we fit mixed-effects models including within-sender-centered positivity as a predictor, alongside condition and random sender intercepts. Within-sender centering removes each individual's baseline positivity level, isolating the contribution of email-by-email tone variation. Within-sender positivity was a strong predictor of email opening (OR=2.05, 95% CI [1.95, 2.16], $z=17.25$, $p<0.001$): each one-point increase in positivity was associated with a doubling of the odds of opening the email. The effect on replying was even larger (OR=3.32, 95% CI [3.08, 3.58], $z=23.64$, $p<0.001$), corresponding to a 232% increase in the odds of receiving a reply (Fig. 1b). These associations were robust after controlling for condition assignment and sender random effects.

Stratifying by recipient type revealed that the positivity–open rate association was significant for internal recipients (OR=1.26, p=0.004) but not for external recipients (OR=1.07, p=0.243), suggesting that playful emotional cues in emails matter most for colleagues within familiar organizational contexts. The positivity–reply rate association was strong and comparable for both internal (OR=1.59, p<0.001) and external recipients (OR=1.60, p<0.001). That the within-recipient-type odds ratios (around 1.6) are smaller than the pooled estimate (3.32) indicates that part of the pooled association operates between contexts: internal emails were both warmer and more frequently engaged with than external ones, and the stratified models remove this between-context gradient to isolate the within-context effect of tone. Descriptively, the relationship followed a monotone gradient across positivity quartiles: emails in the most negative quartile had a 50.2% open rate and 14.0% reply rate, compared with 66.7% and 33.3% in the most positive quartile (Fig. 1c).

**Indirect effects of condition on behavior through positivity**

We tested formal mediation using the Sobel product-of-coefficients method, combining the a-path coefficients (condition → positivity, from the LMM) with the b-path coefficients (positivity → behavior, on the log-odds scale). The indirect effects were highly significant for both outcomes and both active conditions (Table 3). Playful editing produced positive indirect effects on both opening (indirect = +0.049, Sobel z = 7.91, p<0.001) and replying (indirect = 0.081, z = 8.33, p<0.001), whereas professional editing produced negative indirect effects on opening (indirect = −0.030, z = −5.14, p<0.001) and replying (indirect = −0.049, z = −5.25, p<0.001).

The indirect pathway was significant even though the condition had no total effect on behavior. Entering within-sender positivity as a covariate left the already-null direct effects of Playful essentially unchanged (opening OR: 1.039 → 0.990; replying OR: 1.000 → 0.925). AI editing thus influenced behavior only through the tone it produced as it shifted positivity and positivity predicted engagement, but the condition-level differences were too small to surface as a detectable total effect. Tone, not the act of AI editing, was the operative channel.

| Condition | Outcome | a (→pos) | b (pos→) | Indirect (a×b) | Sobel z | p-value |
|---|---|---|---|---|---|---|
| Playful | Opened | 0.068 | 0.718 | 0.049 | 7.91 | <0.001 |
| Playful | Replied | 0.068 | 1.199 | 0.081 | 8.33 | <0.001 |
| Professional | Opened | −0.041 | 0.718 | −0.030 | −5.14 | <0.001 |
| Professional | Replied | −0.041 | 1.199 | −0.049 | −5.25 | <0.001 |

**Table 3. Mediation analysis: indirect effects of condition on behavior through positivity.** a-path: condition → positivity (LMM coefficient). b-path: within-sender centered positivity → outcome (GLMM log-OR). Indirect effect SE estimated via delta method.

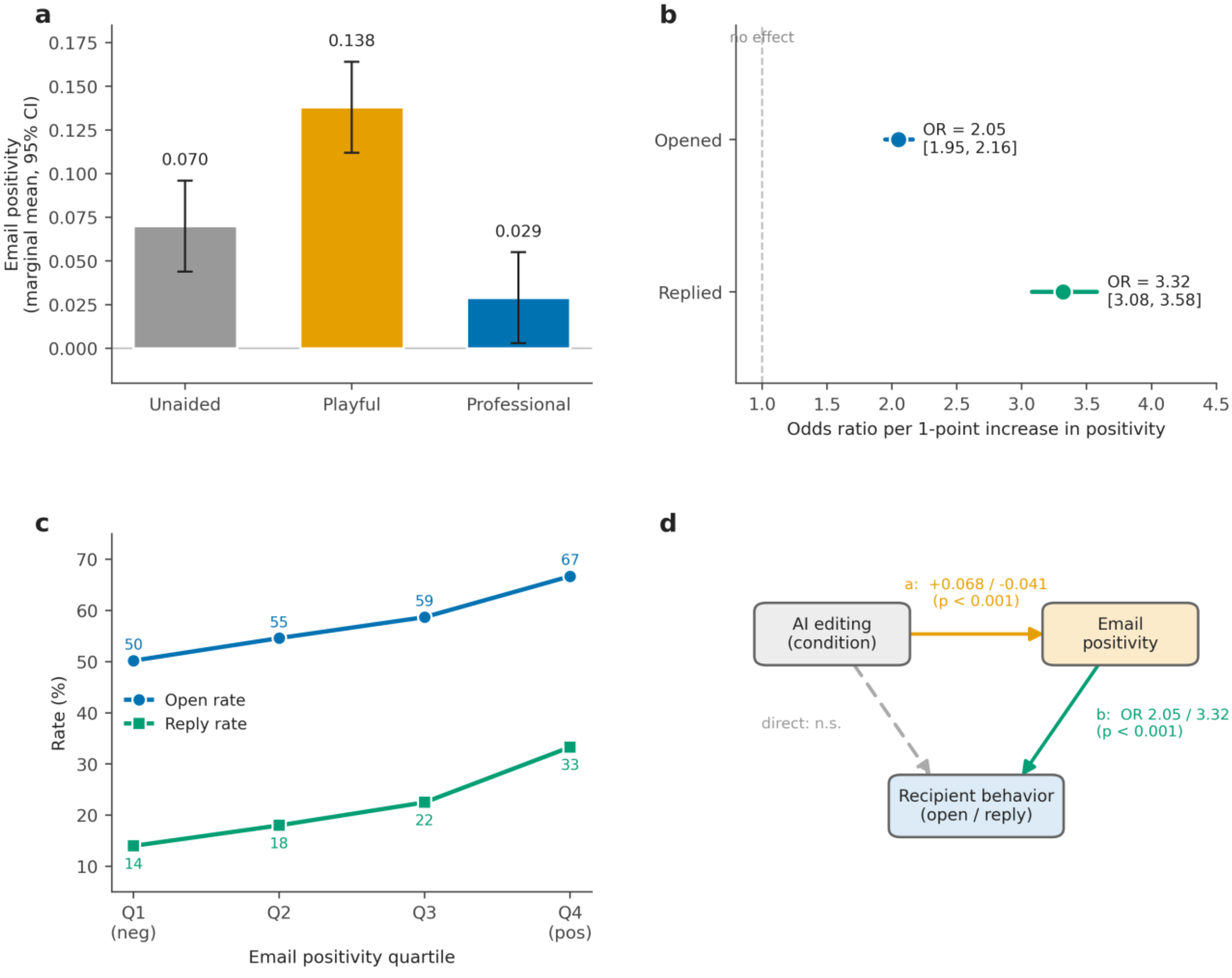

**Figure 1. LLM-assisted email rewriting affects recipient engagement only by shifting emotional tone.** **(a)** Marginal mean positivity scores (±95% CI) by condition. All pairwise contrasts significant after Holm correction ($p<0.001$). **(b)** Association between within-sender centered positivity and behavioral outcomes, controlling for condition. Error bars are 95% CIs for OR estimates. **(c)** Open and reply rates by positivity quartile (Q1=most negative, Q4=most positive). **(d)** Mediation model of condition effects on behavior through email positivity.

## Discussion

Across 16,880 emails sent by 121 employees under randomly assigned conditions, LLM-assisted rewriting reliably shifted the emotional tone of outgoing messages. The tone, not AI involvement, determined whether recipients opened and responded to these messages. Playful editing increased positivity, and professional editing decreased it, yet neither condition directly changed opening, replying, or response speed. The behavioral consequences of AI editing instead ran through a single channel: the positivity of the language recipients actually read. This indirect pathway was significant for both opening and replying, even in the absence of any direct effect of the condition. Our primary hypothesis, that playful editing would broadly increase engagement, was therefore not supported; the effect was real but mediated, and became visible only once we looked past the manipulation to the feature it changed.

Why should positivity matter so much? Positive affect widens attention and motivates approach behavior[15], and the emotions carried in a message act as social information that shapes the receiver's response[13], so a recipient meeting a warmer, more animated message may be readier to engage with its content and to reply. This is consistent with organizational research showing that positive, humorous communication builds rapport and narrows perceived social distance between correspondents[5,6]. What our data add is that the effect operates within the ordinary range of a single sender's writing: email-by-email shifts in positivity, well inside one person's normal variation, were associated with a doubling of the odds of opening (OR=2.05) and more than a tripling of the odds of reply (OR=3.32), both pooled across recipient types. Aggregated across the full range, the gradient was substantial: moving from the least to the most positive quartile of emails corresponded to a 17-percentage-point rise in open rate (50% to 67%) and a 19-point rise in reply rate (14% to 33%).

The absence of any direct behavioral difference between conditions (unaided, playful and professional) is not a failed manipulation but a substantive result. The tone shift itself was large and highly significant. Because recipients were never told whether an email had been AI-edited, the condition could influence them only through features they could perceive, and it did so entirely through tone. This is precisely what the AI-mediated communication framework predicts[7]. AI shapes interpersonal outcomes through the features of the messages it produces, not through its mere presence. The implication generalizes beyond LLMs: any writing intervention will move behavior only to the extent that it changes the message features recipients actually respond to. In this sense, the model was neither a bonus nor a penalty in itself; it was an instrument, and its behavioral footprint was commensurate with the tone shift it produced.

This carries a direct corollary for how such tools are deployed. Because professional-style rewriting reduced positivity, it suppressed engagement through the same mediated pathway that playful rewriting had used to raise it. As general-purpose writing assistants become standard workplace tools[1,2], employees who use them to "make it more professional" may therefore trade warmth for polish[16], and (without intending to) produce messages that read as more correct but invite less response. Formality is not without value; a formal register can convey competence and credibility in some settings[17], and engagement is only one of several outcomes a message may be designed to achieve. The behavioral value of an LLM writing assistant thus depends on the direction in which it shifts the tone, not on whether it improves the text in the abstract, a distinction worth building into the prompts and defaults these tools ship with.

The two recipient types differed in how tone translated into behavior. Playful editing shifted positivity more for external recipients, but for internal recipients, that positivity translated more strongly into email opening. A parsimonious account is expectancy violation[18]: colleagues who share a sender's organizational context hold a well-formed expectation of that sender's usual register, so a playfully atypical email registers as a salient departure and is more likely to be opened. External recipients, lacking that baseline expectation, have no comparable norm to violate, so tone carries less weight at the opening stage. Once a message was opened, however, positivity predicted replying about equally across recipient types (OR≈1.6 in both), suggesting that warmth prompts reciprocation fairly uniformly regardless of prior relationship[13,19].

Several limitations bound these conclusions. The crossover lasted three weeks, so we cannot speak to longer-term dynamics such as recipient habituation, drift in communication norms, or changes in senders' own writing over time. Positivity was captured with an automated polarity score[20,21], which indexes valence but not finer qualities that human recipients likely also register (e.g., humor, authenticity, contextual fit). Because positivity was measured rather than independently manipulated, the tone-to-behavior pathway is correlational, and unobserved features that covary with it (e.g., message length, specificity, topic) cannot be fully ruled out as contributors. We did not collect recipient-side data on whether an email was perceived as AI-edited; whether recipients who detect AI involvement respond differently is an open, increasingly consequential question[22]. Finally, participation was voluntary, which may bias the sample toward employees with stronger communication motivation and limit generalization.

These results establish a tone-mediated link between AI-assisted writing and recipient behavior, but leave open how far the pathway extends across models, prompting strategies, organizational cultures, and channels beyond email. The most pressing extension concerns disclosure. As transparency norms around AI in professional communication take shape[7–9], whether recipients know an email was AI-edited may itself become a primary moderator of the effects we document. Disclosure could add a direct pathway, through trust or authenticity judgments, alongside the indirect, tone-based one identified here. Longitudinal designs that follow organizations as they adopt these tools at scale would reveal whether an individual-level mechanism of this kind aggregates into organization-level change in how well people communicate. Whatever form these tools take, their influence will run through the channel we identify here: the tone of the message that finally reaches a human reader.

## Methods

### Study design

The study used a randomized crossover design in which each participant experienced all three experimental conditions (Unaided, Professional, Playful) in counterbalanced order over three consecutive weeks (one condition per week). Six possible orderings of the three conditions were assigned to subgroups within each company, balancing period and carryover effects across the sample. Subgroups were additionally composed to be demographically comparable, with sex composition, age distribution, and the mix of organizational positions held similar across ordering subgroups. All study procedures were approved by the University of Haifa's Ethics Committee. Informed consent was obtained from all participants prior to data collection.

### Participants and setting

Participants were full-time or part-time employees (age $\geq$ 18) recruited from six companies in Israel, the United Kingdom, Sweden, and Ukraine, all of whom used organizational email daily as part of their regular work. Throughout the study, participants continued to send emails as their work required, so the study imposed no communication quota and observed naturally occurring workplace email behavior. Companies spanned the software, hardware, services, and medical device sectors and ranged in size from 7 to 38 employees (Table S1).

Within each company, interested employees responded to an internal announcement distributed by management.

Participants were asked not to discuss the study with colleagues to limit cross-condition influence within a company. Researchers had no access to employee lists, and participation was entirely voluntary with no occupational consequences for non-participation. Participants were informed they were taking part in a study on "workplace email communication patterns". The specific manipulation (LLM use and condition assignment) was disclosed in a full debriefing letter after data collection ended, in accordance with APA Standard 8.07 on minimal deception. A total of 121 employees (88 males, 33 females; mean age=38.5 years, SD=7.4, range=22–57) across the six companies completed the full three-week protocol and were included in the analysis.

**Experimental conditions**

Each participant experienced three writing conditions, differing only in whether and how an LLM rewrote each drafted email before it was sent: unaided writing, playful LLM rewriting, and professional LLM rewriting.

In the *Unaided* condition, participants composed and sent emails as they normally would, without any AI assistance.

In the *Playful* condition, participants drafted emails independently and then resubmitted them to GPT-5 using a fixed prompt template that instructed the model to rewrite the message in a specified playful style (e.g., pirate-style phrasing, medieval or theatrical voice), while preserving all factual content and intended actions. The specific prompt was: "Transform the following email into a [STYLE] version of the same message. Preserve the exact meaning, all factual details, and all requested actions. Do not introduce new information or omit existing information. Only change the tone, wording, and style. Keep the result readable and suitable for a workplace context. Style: [STYLE]. Email: [EMAIL CONTENT]."

In the *Professional* condition, participants drafted emails independently and then used GPT-5 with a fixed template instructing formal rewriting. The specific prompt was: "Rewrite the following email in a formal and professional workplace tone while preserving its original meaning, factual content, and requested actions. Do not add or remove information. Do not change names, dates, deadlines, or action items. Keep the message clear, concise, and appropriate for professional communication. Email: [EMAIL CONTENT]."

To monitor treatment fidelity, a randomly selected subset of emails in the two LLM conditions was reviewed by a researcher to verify content preservation, appropriate tone, the absence of offensive material, and a length comparable to the original draft (±30 words). All LLM rewriting used OpenAI's GPT-5 via the public API at default temperature; the full prompt templates reproduced above, together with the exact model configuration, are archived in the study repository (see Code availability).

## Measures

*Email open rates* and *reply rates* were recorded as binary indicators per email. Open events were detected via a tracking pixel embedded in outgoing messages that registered the first retrieval event with the study server. A reply receipt was recorded in the sender's email system logs. Time to open and time to reply were computed as elapsed minutes from email delivery to the first open or first reply event.

*Recipient type* was recorded for each email as *internal* (sent to a colleague within the sender's company) or *external* (sent to a recipient outside the company). This distinction was retained throughout the analyses because response norms and communication dynamics may differ between internal and external correspondence.

*Email positivity* was quantified using the *'spacytextblob'* model applied to the full text of each outgoing email, which generates a continuous polarity score (range −1 to +1) representing the degree of positive versus negative emotional tone[20,21]. Scores were computed from the final sent version of each email (i.e., the LLM-edited version in the Playful and Professional conditions).

*Treatment compliance* was assessed post-hoc using the LLM-writing detector proposed by Lazebnik and Rosenfeld[23], which evaluates the probability that a text was generated or substantially edited by an LLM. This check was intended to detect deviations from the assigned condition, in particular emails in the Unaided condition that showed signs of AI assistance despite instructions, and to strengthen confidence in treatment fidelity. The resulting 'LLM detector score' was used in sensitivity analyses comparing intent-to-treat estimates with compliance-restricted samples.

## Data collection and privacy

All emails sent during the study period were extracted from each company's SMTP (Simple Mail Transfer Protocol) provider (e.g., Gmail) and uploaded to a secure server within each company's IT environment. All data processing (open- and reply-event detection, timing, sentiment scoring, and compliance analysis) was performed on this in-company server, and only aggregated, de-identified outcome variables were exported to a central analysis server operated by the authors. No raw email content ever left the companies' systems, preserving the privacy of both the participating employees and their correspondents.

**Statistical analysis**

All analyses were conducted in R (version 4.5.2). The primary analysis followed a within-subject framework exploiting the crossover structure of the design. The unit of analysis was the individual email event. The resulting data were longitudinal and hierarchical, with multiple emails nested within each employee and employees nested within companies. For each email, we retained the sender and company identifiers, the assigned condition, the crossover week, and the recipient type, with participant gender and age retained as covariates. The mixed-effects models described below account for this dependence. The study was not preregistered. Our primary hypotheses concerned the direct effects of condition on the behavioral engagement outcomes (open rate, reply rate, and response speed). The mediation analyses were secondary and exploratory.

*Binary outcomes (emails opened, emails replied)* were modeled using generalized linear mixed-effects models (GLMM; binomial family, logit link) with condition as a fixed effect and random intercepts per sender (‘lme4::glmer’ with ‘bobyqa’ optimizer). Omnibus condition effects were assessed via likelihood ratio tests comparing models with and without condition. Pairwise contrasts were computed using ‘emmeans’ with Holm correction.

*Positivity scores* were modeled using linear mixed-effects models (LMM; REML = FALSE for likelihood-based model comparison) with the same structure. Interactions with recipient type (internal vs. external) were tested by comparing models with and without the interaction term. The interaction model additionally adjusted for gender, age, and company.

*Time-to-event outcomes (time to open, time to reply)* were analyzed using Cox proportional hazards models stratified by sender (‘survival::coxph’), which implement a fully within-subject survival model without distributional assumptions on the baseline hazard. Right-censoring was applied at 14 days for time to open and 40 days for time to reply, corresponding to the observed extremes of the follow-up window. Condition effects were assessed via the omnibus likelihood-ratio test, with hazard ratios reported per condition relative to Unaided baseline.

*Mediation analysis* used within-sender centered positivity (each email's positivity minus the sender's mean across all emails) as the mediator to isolate email-level variation from stable sender-level tendencies. The b-path (centered positivity → outcome) was estimated via GLMM, including condition as a covariate, and was additionally estimated separately within each recipient type (internal, external). Indirect effects were quantified using the Sobel product-of-coefficients method (indirect = a × b; SE estimated via the delta method). Statistical significance was assessed by two-tailed z-tests.

*Treatment-Compliance Sensitivity analyses* included restricting to emails that passed the LLM-detector compliance check (see Supplementary Analysis).

**Data availability**

De-identified, aggregated data supporting the findings of this study are available upon reasonable request to the study PIs (Z.B.Z. and T.L.). Raw email text cannot be shared, as it contains identifiable workplace correspondence; only the derived, de-identified variables used in the analyses can be provided.

**Code availability**.

All analysis code used to reproduce the results is available at https://github.com/zivbz1/Playful-AI-in-Professional-Email. The GPT-5 prompt templates are reproduced verbatim in the Methods.

# Appendix

### Supplementary Analysis: Treatment-Compliance Sensitivity

Assignment to the Playful and Professional conditions did not guarantee that a sent email had in fact been rewritten by the LLM: a participant could submit a draft the model altered only slightly, or send their own text. Our primary analyses are therefore intent-to-treat, analysing every email under its assigned condition regardless of realised compliance. To test whether the absence of a direct condition effect reflects genuine behavioural equivalence rather than a diluted manipulation, we repeated the key analyses on a compliance-restricted (per-protocol) sample. Compliance was operationalised with the LLM-writing detector of Lazebnik and Rosenfeld: an email in the Playful or Professional condition was retained if its detector score was ≥0.50 (indicating genuine LLM editing), and an Unaided email was retained if its score was <0.50 (indicating no LLM editing). Under this criterion the two LLM conditions were fully compliant (Playful 100%, Professional 100%; every email detected as LLM-edited), whereas 34.6% of Unaided emails were excluded for scoring above the detection threshold (Unaided compliance 65.4%); the compliance-restricted sample comprised 14,931 emails. As shown in Table S2, the direct-effect nulls and the tone-mediated indirect effects were essentially unchanged in the compliance-restricted sample, indicating that the null total effect of condition is not an artefact of non-compliance.

**Table S1. Description of the companies involved in the experiments**

Each row describes a company whose employees participated in the field experiment. Company Index is the identifier used to refer to the company throughout the supplementary materials; Country is the company's primary country of operation; Number of Participants is the count of employees enrolled from that company; and Field / Sector denotes the company's industry. The six companies contributed 121 participants in total, spanning software, hardware, medical-device, and IT-services firms across four countries.

| Company Index | Country | Number of Participants | Field / Sector |
|---|---|---|---|
| 1 | Israel | 7 | Software - logistics |
| 2 | Israel | 21 | Software - Music |
| 3 | Israel | 38 | Hardware - Energy |
| 4 | United Kingdom | 15 | Services - IT |
| 5 | Sweden | 22 | Medical Device |
| 6 | Ukraine | 18 | Software - finance |

**Table S2. Treatment-compliance sensitivity analysis**

Each row compares the intent-to-treat estimate (all 16,880 emails, as reported in the main text) with the compliance-restricted (per-protocol) estimate. Direct-effect rows report the likelihood-ratio $\chi^2(2)$ and p; b-path rows report the within-sender positivity odds ratio (per 1-point increase); indirect-effect rows report the Sobel z (all indirect effects $p<0.001$ in the main analysis). Compliance-restricted values were computed by the compliance section of the analysis pipeline.

| Result | Intent-to-treat (N=16,880) | Compliance-restricted |
|---|---|---|
| Direct effect on opening — LRT $\chi^2(2)$, p | $\chi^2$=0.94, p=0.62 | $\chi^2$=0.57, p=0.75 |
| Direct effect on replying — LRT $\chi^2(2)$, p | $\chi^2$=2.34, p=0.31 | $\chi^2$=1.89, p=0.39 |
| Positivity → opening (b-path OR) | 2.05 | 2.06 |
| Positivity → replying (b-path OR) | 3.32 | 3.45 |
| Indirect: Playful → opening (Sobel z) | 7.91 | 7.30 |
| Indirect: Playful → replying (Sobel z) | 8.33 | 7.68 |
| Indirect: Professional → opening (Sobel z) | -5.14 | -4.40 |
| Indirect: Professional → replying (Sobel z) | -5.25 | -4.48 |